\documentclass[10pt,twocolumn,letterpaper]{article}

\usepackage{cvpr}
\usepackage{times}
\usepackage{epsfig}
\usepackage{graphicx}
\usepackage{amsmath}
\usepackage{amssymb}
\usepackage{subfigure}
\usepackage{threeparttable}
\DeclareMathOperator*{\argmin}{argmin}
\DeclareMathOperator*{\argmax}{argmax}
\usepackage{multirow}
\usepackage{makecell}
\usepackage{caption} 

\usepackage[pagebackref=true,breaklinks=true,letterpaper=true,colorlinks,bookmarks=false]{hyperref}

\cvprfinalcopy 

\def\cvprPaperID{11940} 

\begin{document}
\title{Neural Architecture Search with Random Labels}
\author{Xuanyang Zhang \hspace{20pt} Pengfei Hou \hspace{20pt} Xiangyu Zhang\footnotemark[1] \hspace{20pt} Jian Sun \\ 
MEGVII Technology \\
{\tt\small \{zhangxuanyang,houpengfei,zhangxiangyu,sunjian\}@megvii.com}
}

\maketitle
\renewcommand{\thefootnote}{\fnsymbol{footnote}}
\footnotetext[1]{Corresponding author. This work is supported by The National Key Research and Development Program of China (No.2017YFA0700800) and Beijing Academy of Artificial Intelligence (BAAI).} 

\begin{abstract}
In this paper, we investigate a new variant of neural architecture search (NAS) paradigm -- searching with random labels (RLNAS). The task sounds counter-intuitive for most existing NAS algorithms since random label provides few information on the performance of each candidate architecture. Instead, we propose a novel NAS framework based on ease-of-convergence hypothesis, which requires only random labels during searching. The algorithm involves two steps: first, we train a SuperNet using random labels; second, from the SuperNet we extract the sub-network whose weights change most significantly during the training. Extensive experiments are evaluated on multiple datasets (e.g. NAS-Bench-201 and ImageNet) and multiple search spaces (e.g. DARTS-like and MobileNet-like). Very surprisingly, RLNAS achieves comparable or even better results compared with state-of-the-art NAS methods such as PC-DARTS, Single Path One-Shot, even though the counterparts utilize full ground truth labels for searching. We hope our finding could inspire new understandings on the essential of NAS. Code is available at \url{https://github.com/megvii-model/RLNAS}.
\end{abstract}
\section{Introduction}

Recent years \textsl{Neural Architecture Search}~\cite{zoph2016neural, baker2016designing, zoph2018learning, zhong2018practical, zhong2018blockqnn, liu2018progressive, real2019regularized, tan2019mnasnet, chen2019detnas}~(NAS) has received much attention in the community as its superior performances over human-designed architectures on a variety of tasks such as image classification \cite{tan2019mnasnet,tan2019efficientnet,guo2019single}, object detection \cite{chen2019detnas,ghiasi2019fpn} and semantic segmentation \cite{liu2019auto}. In general, most existing NAS frameworks can be summarized as a \emph{nested bilevel optimization}, formulated as follows:
\begin{equation}
a^{\star} = \mathop{\argmax}_{a \in \mathcal{A}} \text{Score} \left(a, \mathbf{W}_a^{\star}\right) 
\label{eq0}
\end{equation}
\begin{equation}
\mbox{s.t.}\quad \mathbf{W}_a^{\star} = \mathop{\argmin}_{\bf{W}} \mathcal{L} \left(a, \mathbf{W}\right),
\label{eq1}
\end{equation}
where $a$ is a candidate architecture with weights $\mathbf{W}_{a}$ sampled from the search space $\mathcal{A}$; $\mathcal{L}(\cdot)$ represents the \emph{training loss}; $\text{Score}(\cdot)$ means the performance indicator (e.g. accuracy in supervised NAS algorithms or \emph{pretext task} scores in unsupervised NAS frameworks \cite{liu2020labels}) evaluated on the \emph{validation set}. Briefly speaking, the NAS paradigm aims to search for the architecture which obtains the best validation performance, thus we name it \emph{performance-based NAS} in the remaining text.

Despite the great success, to understand why and how \emph{performance-based NAS} works is still an open question. Especially, the mechanism how NAS algorithms discover good architectures from the huge search space is well worth study. 
A recent literature \cite{Shu2020Understanding} analyzes the searching results under cell-based search spaces and reveals that existing performance-based methods tend to favor architectures with fast convergence. Although Shu \textit{et al.}~\cite{Shu2020Understanding} further empirically find that architectures with fast convergence can not achieve the highest generalization performance, the fast convergence connection pattern still implies that there may exist high correlations between architectures with fast convergence and the ones with high performance (named \emph{ease-of-convergence hypothesis} for short). Inspired by the hypothesis, we propose an alternative NAS paradigm, \emph{convergence-based NAS}, as follows: 
\begin{equation}
a^{\star} = \mathop{\argmax}_{a \in \mathcal{A}} \text{Convergence} \left(a, \mathbf{W}_{a}^{\star}\right) 
\label{eq2}
\end{equation}
\begin{equation}
\mbox{s.t.}\quad \mathbf{W}_{a}^{\star} = \mathop{\argmin}_{\mathbf{W}} \mathcal{L} \left(a, \mathbf{W}\right),
\label{eq3}
\end{equation}
where $\text{Convergence}(\cdot)$ is a certain indicator to measure the speed of convergence; other notations follow the same definitions as in Eq.~\ref{eq0},~\ref{eq1}. 

In this paper we mainly investigate \emph{convergence-based NAS} frameworks, which is rarely \emph{explicitly} explored in previous works to our knowledge. First of all, we study the role of \emph{labels} in both frameworks. In \emph{performance-based NAS}, we notice that \emph{feasible labels} are critical in both search steps:~for Eq.~\ref{eq0} step, since we need to select the architecture with the highest validation performance, reasonable labels such as ground truths or at least carefully-designed \emph{pretext task} (e.g. rotation prediction \cite{gidaris2018unsupervised}) labels in unsupervised NAS~\cite{liu2020labels} are required for evaluation. For Eq.~\ref{eq1} step such corresponding labels are also necessary in the training set to optimize the weights. While in \emph{convergence-based NAS}, Eq.~\ref{eq2}
only depends on a metric to estimate the convergence speed, which is free of labels. Though the optimization in Eq.~\ref{eq3} still needs labels, the purpose of the training is just to provide the evidence for the benchmark in Eq.~\ref{eq2} rather than accurate representations. So, we conclude that in convergence-based NAS the requirement of labels is much weaker than that in performance-based NAS. 

The observation motivates us to take a further step: in convergence-based NAS, \emph{can we use only random labels for search, instead of any feasible labels like ground truths or pretext task labels entirely?} To demonstrate it, we propose a novel convergence-based NAS framework, called \emph{Random Label NAS (RLNAS)}, which only requires random labels to search. RLNAS follows the paradigm of Eq.~\ref{eq2},~\ref{eq3}. In Eq.~\ref{eq3} step, random labels are adopted to optimize the weight for each sampled architecture $a$; while in Eq.~\ref{eq2} step, a customized \emph{angle} metric \cite{hu2020angle} is introduced to measure the distance between trained and initialized weights, which estimates the convergence speed of the corresponding architecture. To speed up the search procedure, RLNAS further utilizes the mechanism of \emph{One-Shot NAS} \cite{bender2018understanding,guo2019single} to decouple the nested optimization of Eq.~\ref{eq2} and Eq.~\ref{eq3} into a \emph{two-step} pipeline: first training a \emph{SuperNet} with random labels, then extracting the sub-network with the fastest convergence speed from the SuperNet using evolutionary search. 

We evaluate our RLNAS in popular search spaces like NAS-Bench-201~\cite{dong2020bench}, DARTS~\cite{liu2018darts} and MobileNet-like search space~\cite{cai2018proxylessnas}. Very surprisingly, though RLNAS does not use any feasible labels, it still achieves comparable or even better performances on multiple benchmarks than many supervised/unsupervised methods, including state-of-the-art NAS frameworks such as \emph{PC-DARTS} \cite{xu2019pc}, \emph{Single-Path One-Shot} \cite{guo2019single}, \emph{FairDARTS} \cite{chu2019fair}, \emph{FBNet} \cite{wu2019fbnet} and \emph{UnNAS} \cite{liu2020labels}. Moreover, networks discovered by RLNAS are also demonstrated to transfer well in the downstream tasks such as object detection and semantic segmentation.   

In conclusion, the major contribution of the paper is that we propose a new convergence-based NAS framework \emph{RLNAS}, which makes it possible to search with only random labels. We believe the potential of RLNAS may includes:

\vspace{-1em}
\paragraph{A simple but stronger baseline.} Compared with the widely used \emph{random search} \cite{li2020random} baseline, RLNAS is much more powerful, which can provide a stricter validation for future NAS algorithms.   


\vspace{-1em}
\paragraph{Inspiring new understandings on NAS.} Since the performance of RLNAS
is as good as many supervised NAS frameworks, on one hand, it further validates the effectiveness of \emph{ease-of-convergence hypothesis}. On the other hand, however, it suggests that the ground truth labels or NAS on specified tasks do not help much for current NAS algorithms, which implies that architectures found by existing NAS methods may still be suboptimal.

\section{Related Work}
\paragraph{Supervised Neural Architecture Search.} Supervised neural architecture search~(NAS) paradigm is the mainstream NAS setting. Looking back the development history, supervised NAS can be divided into two categories hierarchically: \emph{nested NAS} and \emph{weight-sharing NAS} from the perspective of search efficiency. In the early stage, \emph{nested} \textsl{NAS}~\cite{zoph2016neural, baker2016designing, zoph2018learning, zhong2018practical, zhong2018blockqnn, liu2018progressive, real2019regularized, tan2019mnasnet} trains candidate architectures from scratch and update controller with corresponding performance feedbacks iteratively. However, \emph{nested NAS} works at the cost of a surge in computation, e.g. NASNet~\cite{zoph2018learning} costs about 1350--1800 GPU days. ENAS~\cite{pham2018efficient} observes the computation bottleneck of \emph{nested NAS} and forces all candidate architectures to share weights. ENAS takes $1000\times$ less computation cost than \emph{nested NAS}~\cite{pham2018efficient} and proposes a new NAS paradigm named \emph{weight-sharing NAS}. A large number of literature~\cite{liu2018darts, chen2019progressive, xu2019pc, bender2018understanding,brock2017smash, cai2018proxylessnas, guo2019single} follow the \emph{weight-sharing} strategy due to the superiority of search efficiency. This work is also carried out under the \emph{weight-sharing} strategy. Unlike most \emph{weight-sharing} approaches, we are not focusing on the improvement of search efficiency.

According to different optimization steps, \emph{weight-sharing} approaches can be further divided into two categories: the one joint step optimization approach named \emph{gradient-based NAS}~\cite{liu2018darts, chen2019progressive, xu2019pc}) and the two sequential steps optimization approach named \emph{One-Shot NAS}~\cite{bender2018understanding,brock2017smash, cai2018proxylessnas, guo2019single}). The \emph{gradient-based NAS} relaxes discrete search space into a continuous one with architecture parameters, which are optimized with end-to-end paradigms. Because of the non-differentiable characteristic of angle, we follow the mechanism of \emph{One-Shot NAS} to study \emph{convergence-based NAS}.  
\paragraph{Unsupervised Neural Architecture Search.} Recently, unsupervised learning~\cite{He_2020_CVPR, chen2020simple, grill2020bootstrap} has received much attention, and the unsupervised paradigm has also appeared in the field of NAS. \cite{yan2020does} used unsupervised architecture representation in the latent space to better distinguish network architectures with different performance. UnNAS~\cite{liu2020labels} introduces unsupervised methods~\cite{gidaris2018unsupervised, noroozi2016unsupervised, zhang2016colorful} to \emph{weight-sharing NAS} in order to ablate the role of labels. Although UnNAS does not use the labels of the target dataset, the labels like \emph{rotation category}, etc on the pretext tasks are still exploited. UnNAS shows that \emph{weight-sharing NAS} can still work with the absence of ground truth labels, but it is hard to conclude that labels are completely unnecessary. Different from unsupervised learning, which requires representation, unsupervised NAS focuses on architectures. Therefore, random labels are introduced in this paper, which completely detach from prior supervision information and help us thoroughly ablate the impact of labels on NAS.

\paragraph{Model Evaluation Metrics.} \cite{mellor2020neural, anonymous2021neural} develop \emph{training-free NAS} which means searching directly at initialization without involving any training. They focus on investigating training-free model evaluation metrics to rank candidate architectures. \cite{mellor2020neural} uses the correlation between input Jacobian to indicate model performance. \cite{anonymous2021neural} uses the combination of NTKs and linear regions in input space to measure the architecture trainability and expressivity. Although \emph{training-free NAS} has much higher search efficiency, there is still a performance gap compared with well-trained \emph{weight-sharing NAS}. ABS~\cite{hu2020angle} introduces angle metric to indicate model performance and mainly focuses on search space shrinking. Different from ABS, we directly search architectures with angle metric. 

\section{Methodology}

As mentioned in the introduction, in order to utilize the mechanism of Oner-Shot NAS, we first briefly review Single Path One-Shot~(SPOS)~\cite{guo2019single} as preliminary. Based on SPOS framework, we then put forward our approach Random Label NAS~(RLNAS).  
\subsection{Preliminary:~SPOS} 
\label{methodology.preliminaries}

SPOS is one of the One-Shot approaches, which decouple the NAS optimization problem into two sequential steps: firstly train SuperNet, and then search architectures. Different from other One-Shot approaches, SPOS further decouples weights of candidate architectures by training SuperNet stochastically. Specifically, SPOS regards a candidate architecture in SuperNet as a single path and uniformly activates a single path to optimize corresponding weights in each iteration. Thus, the SuperNet training step can be expressed as:

\begin{equation}
\mathbf{W}_{a}^{\star} = \mathop{\argmin}_{\mathbf{W}} \mathbb{E}_{a \sim \varGamma(\mathcal{A}) }
\mathcal{L} \left(a, \mathbf{W}\right),
\label{eq4}
\end{equation}
where $\mathcal{L}$ means objective function optimized on training dataset with ground truth labels and $\varGamma(\mathcal{A})$ is a uniform distribution of $a \in \mathcal{A}$. 

After SuperNet trained to convergence, SPOS performs architecture search as:
\begin{equation}
a^{\star} = \mathop{\argmax}_{a \in \mathcal{A}} \text{ACC}_{val} \left(a, \mathbf{W}_{a}^{\star}\right).
\label{eq5}
\end{equation}
SPOS implements Eq.~\ref{eq5} by utilizing an evolution algorithm to search architectures. With initialized population, SPOS conducts crossover and mutation to generate new candidate architectures and uses validation accuracy as fitness to keep candidate architectures with top performance. Repeat this way until the evolution algorithm converges to the optimal architecture.

\subsection{Our approach: Random Label NAS~(RLNAS)}
\label{methodology.approach}

The combination of two decoupled optimization steps, SuperNet structure consisting of single paths and evolution search, makes SPOS simple but flexible. Following the mechanism of SPOS, we decouple the \emph{convergence-based} optimization of Eq.~\ref{eq2} and Eq.~\ref{eq3} into the following two steps. 

Firstly, SuperNet is trained with random labels:
\begin{equation}
\mathbf{W}_{a}^{\star} = \mathop{\argmin}_{\mathbf{W}} \mathbb{E}_{a \sim \varGamma(\mathcal{A}) }
\mathcal{L} \left(a, \mathbf{W}, R\right),
\label{eq6}
\end{equation} 
where $R$ represents random labels; other notations follow the same definitions as in Eq.~\ref{eq4}.

Secondly, evolution algorithm with convergence-based metric $\text{Convergence}(\cdot)$ as fitness searches the optimal architecture from SuperNet:
\begin{equation}
a^{\star} = \mathop{\argmax}_{a \in \mathcal{A}} \text{Convergence} \left(a, \mathbf{W}_{a}^{\star}\right).
\label{eq7}
\end{equation}

In the next section, we introduce the mechanism of generating random labels in Sec.~\ref{methodology.random_label} and use an angle-based metric as $\text{Convergence}(\cdot)$ to estimate model convergence speed in Sec.~\ref{methodology.angle}.

\subsubsection{Random Labels Mechanism}
\label{methodology.random_label}
In representation learning field, deep neural networks~(DNNs) have the capacity to fit dataset with partial random labels~\cite{zhang2016understanding}. Further more, \cite{maennel2020neural} tries to understand what DNNs learn when trained on natural images with entirely random labels and experimentally demonstrates that pre-training on purely random labels can accelerate the training of downstream tasks under certain conditions. For NAS field, although we pursue the optimal model architecture rather than model representation in search phase, model representation is still involved in the performance-based NAS. However, it is still an open question can neural architecture search work within random labels setting. In the view of this, we try to study the impact of random labels on NAS optimization problem.

At first, we introduce the mechanism of generating random labels. To be specific, random labels obey the discrete uniform distribution and the number of discrete variable is equal to the image category of dataset in default~(other possible methods are discussed in Sec.~\ref{sec.ablation.study}). Random labels corresponding to different images are sampled in data pre-processing procedure and these image-label pairs will not change during the whole model optimization process. 

\subsubsection{Angle-based Model Evaluation Metric}
\label{methodology.angle}

Recently,~\cite{Shu2020Understanding} found out that searched architectures by NAS algorithms share the same pattern of fast convergence. With this rule as a breach, we try to design model evaluation metrics from the perspective of model convergence. \cite{carbonnelle2018layer} firstly measure the convergence of a stand-alone trained model with a angle-based metric. The metric is defined as the angle between initial model wights and trained ones. ABS~\cite{hu2020angle} introduces this metric into the NAS community and uses it to shrink the search space progressively. Different from ABS, we focus on the optimization problem with random labels and adopt angle-based metric to directly search architectures rather than shrink search space. Prior to extend angle to guide architecture search, we first review angle metric in ABS~\cite{hu2020angle}.

\paragraph{Review Angle Metric in ABS.}
SuperNet is represented as a directed acyclic graph~(DAG) denoted as $\mathcal{A}(\mathbf{O},\mathbf{E})$, where $\mathbf{O}$ is the set of feature nodes and $\mathbf{E}$ is the set of connections~(each connection is instantiated as an alternative operation) between two feature nodes. ABS defines $\mathcal{A}(\mathbf{O},\mathbf{E})$ with the only input node $O_{in}$ and the only output node $O_{out}$. A candidate architecture is sampled from SuperNet and it is represented as $a(\mathbf{O},\widetilde{\mathbf{E}})$. The candidate architecture has the same feature nodes $\mathbf{O}$ as SuperNet but subset edges $\widetilde{\mathbf{E}} \in \mathbf{E}$. ABS uses a weight vector $\boldsymbol{V}(a, \mathbf{W})$ to represent a model and constructs $\boldsymbol{V}(a, \mathbf{W})$ by concatenating the weights of all paths from $O_{in}$ to $O_{out}$. The distance between the initialized candidate architecture whose weights is $\mathbf{W_{0}}$ and the trained one with weights $\mathbf{W_{t}}$ is:  
\begin{equation}
    \text{Angle}(a) = \arccos{(\frac{<\boldsymbol{V}(a, \mathbf{W}_{0}),\boldsymbol{V}(a,\mathbf{W}_{t})>}{\left\|\boldsymbol{V}(a,\mathbf{W}_{0})\right\|_{2} \cdot \left\|\boldsymbol{V}(a, \mathbf{W}_{t})\right\|_{2}})}.
    \label{eq9}
\end{equation}

\paragraph{Extensive Representation of Weight Vector.} As above discussed, ABS define the SuperNet with just one input node and one output node. However, for some search spaces, they consist of cell structures with multiple input nodes and outputs nodes. For example, each cell in DARTS has two input nodes and the output node of each cell consists outputs of all intermediate nodes by concatenation, which motivates us to consider all intermediate nodes as output nodes for the identification of architecture topology. In general, we redefine weight vector $\boldsymbol{V}(a, \mathbf{W})$ by concatenating the weights of all paths from $\mathbf{O}_{in}$ to $\mathbf{O}_{out}$. 

\paragraph{Parameterize Non-weight Operations.}So as to resolve the conflict among candidate architectures with the same learnable weights, ABS parameterizes non-weight operations~('pool', 'skip-connect' and 'none'). The 'pool' operation~(both 'average pool' and 'max pool') is assigned with a fixed tensor with dimension $[O,C,K,K]$~($O$ and $C$ represent output channels and input channels respectively, $K$ is the kernel size) and all elements are $1/K^{2}$. Different from ABS assign 'skip-connect' with empty vector, we propose an alternative parametric method, which assigns identity tensor with dimension $[O,C,1,1]$ to the 'skip-connect' operation. We adjust parametric methods for different search spaces, e.g., empty weights and identity tensor are assigned to 'skip-connect' in NAS-Bench-201 and DARTS or MobileNet-like search space respectively. The reason for the difference may be related to the complexity of the search space. The 'none' operation need not to be parameterized as ABS and it determines the number of paths that make up the weights vector~$\boldsymbol{V}$. If there is a 'none' in a path, then weights of operations in this path will not involved in angle calculation.

\section{Experiments}

\subsection{Search Space and Training Setting}
We analyze and evaluate RLNAS on three existing popular search spaces: NAS-Bench-201~\cite{dong2020bench}, DARTS ~\cite{liu2018darts} and MobileNet-like search space~\cite{cai2018proxylessnas}. 
\paragraph{NAS-Bench-201.} There are 6 edges in each cell and each edge has 5 alternative operations. Because of repeated stacking, NAS-Bench-201 consists of 15625 candidate architectures and  provides the real performance for each architecture. We adopt the same training setting for SuperNet in a single GPU across CIFAR-10~\cite{krizhevsky2009learning} CIFAR-100~\cite{krizhevsky2009learning} and ImageNet16-120~\cite{chrabaszcz2017downsampled}. We train the SuperNet 250 epochs with mini-batch 64. We use SGD to optimize weights with momentum 0.9 and weight decay $5e^{-4}$. The learning rate follows cosine schedule from initial 0.025 annealed to 0.001. In evolution phase, we use population size 100, max iterations 20 and keep top-30 architectures in each iteration. All experiment results on NAS-Bench-201 are obtained in three independent runs with different random seeds. 

\begin{table*}[htbp]
  \centering 
  \footnotesize
  \begin{tabular}{l|c|c|c|c|c|c|c|c} 
  \hline 
  \multirow{2}*{Method} & \multicolumn{2}{c}{Configurations} & \multicolumn{2}{|c}{CIFAR-10~(\%)} & \multicolumn{2}{|c}{CIFAR-100~(\%)}  & \multicolumn{2}{|c}{ImageNet16-120~(\%)}  \\
  \cline{2-9}
      ~  & Label type & Performance indicator & valid acc & test acc & \multicolumn{1}{c|}{valid acc} & \multicolumn{1}{c|}{test acc} & valid acc & test acc \\
  \hline\hline
  A~(SPOS)\tnote{$\dagger$} & ground truth label & validation accuracy & 88.49 & 92.11 & 66.51 & 66.89 & 40.16 & 40.80 \\ 
  \hline
  B & ground truth label & angle & \textbf{90.20} & \textbf{93.76} & 70.71 & \textbf{71.11} & 40.78 & 41.44  \\
  \hline
  C & random label & validation accuracy & 76.47 & 80.60 & 52.48 & 52.84 & 29.58 & 28.37 \\
  \hline
  D~(RLNAS)\tnote{$\ddagger$} & random label & angle & 89.94 & 93.45 & \textbf{70.98} & 70.71 & \textbf{43.86} & \textbf{43.70} \\ 
  \hline 
  \end{tabular}

  \caption{ Search performance on NAS-Bench-201 across CIFAR-10, CIFAR-100 and ImageNet16-120.} 
  \label{tab.1}
\end{table*}

\paragraph{DARTS.}Different from vanilla DARTS~\cite{liu2018darts}, each intermediate node only samples two operations among alternative operations~(except 'none') from its all preceding nodes in SuperNet training phase. We train the SuperNet with 8 cell on CIFAR-10 for 250 epochs and other training settings keep the same as DARTS~\cite{liu2018darts}. We also train 14 cell SuperNet with initial channel 48 on ImageNet. We use 8 GPUs to train SuperNet 50 epochs with mini-batch 512. SGD with momentum 0.9 and weight decay $4e^{-5}$ is adopted to optimize weights. The cosine learning rate schedules from 0.1 to $5e^{-4}$. We use the same evolution hyper-parameters as Single Path One-Shot~(SPOS)~\cite{guo2019single}. As for model evaluation phase~(retrain searched architecture), we follow the training setting as PC-DARTS~\cite{xu2019pc} on ImageNet. 

\paragraph{MobileNet.}The MobileNet-like search space proposed in ProxylessNAS~\cite{cai2018proxylessnas} is adopted in this paper. The SuperNet contains 21 choice blocks and each block has 7 alternatives: 6 MobileNet blocks~(combination of kernel size 
\{3,5,7\} and expand ratio \{3,6\}) and 'skip-connect'. We keep the same experiment setting for both search phase and evaluation phase as SPOS~\cite{guo2019single}.

\subsection{Searching Results}
\subsubsection{NAS-Bench-201 Experiment Results}
\label{sec.4.2.1}

\paragraph{Search performance.} For NAS-Bench-201 search space, experiments are conducted on three datasets: CIFAR-10, CIFAR-100 and ImageNet16-120. Different from other literature only search on CIFAR-10 and look up real performance of the found architecture on various test dataset~(e.g., test accuracy on CIFAR-100 or ImageNet16-120), we actually train SuperNet on different target datasets and search architectures with the unique SuperNet. Firstly, we construct SuperNet based NAS-Bench-201 search space and train the SuperNet by uniform sampling strategy~\cite{guo2019single} with ground truth labels or random labels. Then, angle or validation accuracy is regarded as fitness to perform evolution search. According to different method configurations, there are total four possible methods as described in Table~\ref{tab.1}. For simplity, we denoted they as method A, B, C and D respectively. In particular, 
method A and D correspond to SPOS and RLNAS. The search performance on three datasets are reported in Table~\ref{tab.1}.  We first compare method C and D within the random label setting, and find that angle surpasses validation accuracy with a large margin. Similar results can also be observed under the ground truth label setting, but the margin between method A and B is not such large. This suggests that angle can evaluate models more accurately than validation accuracy. Further more, in the case where angle is used as the metric, even if random labels are used, RLNAS obtains comparable accuracy on CIFAR-10 and CIFAR-100 and even outperforms method B by 1.26\% test accuracy on ImageNet16-120.
\paragraph{Ranking correlation.}
   
In addition to the analysis of top architectures as Table~\ref{tab.1}, we further conduct rank correlation analysis. The first step is also to train SuperNet with ground truth labels or random labels. Secondly, we traverse the whole NAS-Bench-201 search space and rank them with different model evaluation metrics independently. We treat the rank based on real performance provided by NAS-Bench-201 as the ground truth rank. At last, we compute the Kendall's Tau~\cite{kendall1938new,yu2019evaluating,chu2019fairnas, hu2020angle} between the rank based on the model evaluation metric and the ground truth rank to evaluate the ranking correlation. We compare angle and validation accuracy as model evaluation metric in both ground truth label and random label setting across three datasets. The ranking correlation results are shown in Table~\ref{tab.2}. The results on different datasets show the consistent order of ranking correlation: C$<$A$<$D$<$B. It should be noted that the rank obtained by validation accuracy in the case of random labels has almost no correlation with the ground truth rank. To our surprise, angle still has the ranking correlation around 0.5 under the random label setting, which even exceeds validation accuracy in ground truth label case.

\begin{table}[htbp]
   \centering 
   \footnotesize
   \begin{tabular}{l|c|c|c} 
   \hline 
   \multirow{1}*{$\text{Method}^\dagger$} & {CIFAR-10} & {CIFAR-100}  & {ImageNet16-120}  \\
   \hline\hline
   A~(SPOS) & 0.4239  &    0.4832  &    0.4322   \\ 
   \hline
   B & \textbf{0.6671}  &    \textbf{0.6942}  &   \textbf{0.6342}   \\
   \hline
   C & 0.0874  & \multirow{1}{1.25cm}{$-$0.0195}  & \multirow{1}{1.25cm}{$-$0.0262}  \\
   \hline
   D~(RLNAS) & 0.5059  & 0.5097   & 0.4716   \\ 
   \hline 
   \end{tabular}

   \caption{Ranking correlation on NAS-Bench-201. $^\dagger$ refer to Table~\ref{tab.1} for detailed method configurations.} 
   \label{tab.2}
\end{table}


\begin{table*}[htbp]
  \centering 
  \footnotesize
  \begin{tabular}{c|l|c|c|c|c} 
  \hline 
  Search type & Method & Params~(M) & FLOPs~(M)&Top-1~($\%$) &Top-5~($\%$) \\
  \hline\hline
  \multirow{7}*{CIFAR-10}& DARTS~\cite{liu2018darts}~(
 \textit{sup.})   & 4.7 & 574 & 73.3 &  91.3  \\
  ~& SPOS~\cite{guo2019single}~(\textit{sup, our impl.})  & 4.3 & 471 & 73.7 & 91.6 \\ 
  ~& PC-DARTS~\cite{xu2019pc}~(\textit{sup.}) & 5.3 & 586 & 74.9 & 92.2  \\
  ~& FairDARTS-B~\cite{chu2019fair}~(\textit{sup.}) & 4.8 & 541 & 75.1 & 92.5 \\
  ~& P-DARTS~\cite{chen2019progressive}~(\textit{sup.})  & 4.9 & 557 & 75.6 & 92.6 \\
  \cline{2-6}
  ~& RLNAS~(\textit{unsup.})  & 5.7 & 629 & \textbf{76.0} & \textbf{92.9} \\ 
  ~& $\text{RLNAS}^\blacktriangledown$~(\textit{unsup.}) & 5.3 & 581 & 75.6 & 92.5 \\ 
  \hline
  \multirow{4}*{ImageNet} & SPOS~\cite{guo2019single}~(\textit{sup, our impl.})   & 4.6 & 512 & 74.5 & 92.1 \\ 
  ~ & $\text{NAS-DARTS}^{\dagger}$~\cite{liu2020labels}~(\textit{sup.})  & 5.3 & 582  & \textbf{76.0} & 92.7 \\
  ~ & PC-DARTS~\cite{xu2019pc}~(\textit{sup.})    & 5.3 & 597 & 75.8 & 92.7 \\
  \cline{2-6}
  ~ & RLNAS~(\textit{unsup.}) & 5.5 & 597 & 75.9 & \textbf{92.9} \\ 
  \hline 
  \end{tabular}
   
  \caption{DARTS search space results: comparison of the SOTA methods on ImageNet. There are two search types of methods and the results of the first block and the second block are searched on CIFAR-10 and ImageNet respectively. $^\blacktriangledown$ FLOPs of the searched architecture is scaled down within 600M by adjusting initial channels from 48 to 46. $^\dagger$ retrain NAS-DARTS reported in UnNAS~\cite{liu2020labels} as PC-DARTS~\cite{xu2019pc}.} 
  \label{tab.3}
\end{table*}

\subsubsection{DARTS Search Space Results}
We conduct two types of experiments in DARTS search space:~search architectures with 8 cells on CIFAR-10, then transfer to ImageNet and search architectures with 14 cells on ImageNet directly. For experiment conducted on CIFAR-10, the training dataset is divided into two subsets with equal size, one of which is used to train the SuperNet, and the other is used as the validation dataset to evaluate model performance in the search phase. As for experiments searched on ImageNet, 50K images are separated from the original training dataset as validation and the rest images are used as the new training dataset. 
\paragraph{Search architectures on CIFAR-10.}
We first analyze the search performance on CIFAR-10 dataset in Table~\ref{tab.3}. RLNAS embodies strong generalization ability when transfering searched architecture from CIFAR-10 to ImageNet. As shown in the first block of Table~\ref{tab.3}, RLNAS has reached 76.0\% top-1 accuracy, even obtains 75.6\% within 600M FLOPs constrain.

\paragraph{Search architectures on ImageNet.}
After demonstrating the transferring ability of RLNAS among classification tasks, we further verify the efficacy of our method by directly searching on ImageNet. To our best knowledge, it is the first time to train SuperNet with \textbf{14 cells} in DARTS search space without any SuperNet structure modification or complicated techniques. After SuperNet training, we search candidate architectures with 600M FLOPs constrain. The searching results are shown in the second block of Table~\ref{tab.3} and RLNAS obtains 75.9\%. Compared with the results found on CIFAR-10, the performance of RLNAS is further improved by 0.3\%, which indicates that narrowing the gap between the training setting~(both dataset and SuperNet structure) of the search phase and the one in the evaluation phase is helpful for architecture search.

\paragraph{Comparison with UnNAS.}Further, we compare our method with UnNAS~\cite{liu2020labels} which also search architectures directly on ImageNet-1K with three pretext tasks~\cite{gidaris2018unsupervised, noroozi2016unsupervised, zhang2016colorful}. For fair comparisons with UnNAS, we have no FLOPs limit in the search phase, but after the search is completed, we limit the FLOPS within 600M by scaling the initial channels from 48 to 42. Simultaneously, we retrain the three architectures reported as UnNAS~\cite{liu2020labels} with the same training setting as PC-DARTS~\cite{xu2019pc}. Table~\ref{tab.4} shows that our method obtains high performance with 76.7\% and 75.9\% within 600M FLOPs constrain, which is comparable with UnNAS with jigsaw task and competitive to results obtained by the other two pretext tasks.

\begin{table}[htbp]
   \centering 
   \footnotesize
   \begin{tabular}{l|c|c|c|c} 
   \hline 
   Method & \makecell[c]{Params \\ (M)} & \makecell[c]{FLOPs \\ (M)}& \makecell[c]{Top-1 \\ ($\%$)} & \makecell[c]{Top-5 \\ ($\%$)}   \\
   \hline\hline
   UnNAS~\cite{liu2020labels}~(\textit{rotation task.})     & 5.1 & 552  & 75.8 & 92.6  \\
   UnNAS~\cite{liu2020labels}~(\textit{color task.})     & 5.3 & 587  & 75.5 & 92.6  \\
   UnNAS~\cite{liu2020labels}~(\textit{jigsaw task.})     & 5.2 & 560  & 76.2 & 92.8  \\
   \hline
   RLNAS~(\textit{random label.}) & 6.6 & 724 & \textbf{76.7} & \textbf{93.1} \\
   $\text{RLNAS}^\blacktriangledown$~(\textit{random label.}) & 5.2 & 561 & 75.9 & 92.8 \\ 
   \hline 
   \end{tabular}

   \caption{DARTS search space results: comparison with UnNAS on ImageNet. The architectures of UnNAS based on three pretext tasks are provided in \cite{liu2020labels} and we retrain them as PC-DARTS training setting~\cite{xu2019pc}.$^\blacktriangledown$ FLOPs of the searched architecture is scaled down within $600$M by adjusting initial channels from 48 to 42. } 
   \label{tab.4}
\end{table}

\subsubsection{MobileNet-like Search Space Results.}

To verify the versatility of our method, we further conduct experiments in the MobileNet-like search space. We train SuperNet with 120 epochs on ImageNet as \cite{guo2019single}. In the search phase, we limit model FLOPs within 475M so as to make fair comparisons with other methods. Results are summarized in Table~\ref{tab.5}. RLNAS obtains 75.6\% top-1 accuracy. Compared with other SOTA methods, our method even outperforms with a slight margin, which verify that our strategy does not overfit to any search space and can achieve effective results generally.

\begin{table}[htbp]
   \centering 
   \footnotesize
   \begin{threeparttable}
   \begin{tabular}{l|c|c|c|c} 
   \hline 
   Method & \makecell[c]{Params\\(M)} & \makecell[c]{FLOPs\\(M)}& \makecell[c]{Top-1~\\(\%)} & \makecell[c]{Top-5 \\ (\%)} \\
   \hline\hline
   FairNAS-A~\cite{chu2019fairnas}~(\textit{sup.})  & 4.6 & 388  & 75.3 & 92.4  \\
   FBNet-C~\cite{wu2019fbnet}~(\textit{sup.})       & 4.4 & 375 & 74.9 & 92.1  \\
   Proxyless~(GPU)~\cite{cai2018proxylessnas}~(\textit{sup.})  & 7.0 & 457 & 75.1 & 92.5   \\
   FairDARTS-D~\cite{chu2019fair}~(\textit{sup.})& 4.3 & 440 & \textbf{75.6} & \textbf{92.6} \\
   SPOS~\cite{guo2019single}~(\textit{sup.})  & 5.4 & 472 & 74.8 & - \\ 
   \hline
   RLNAS~(\textit{unsup.}) & 5.3 & 473 & \textbf{75.6} & \textbf{92.6}  \\ 
   \hline 
   \end{tabular}
   
   \end{threeparttable}
   \caption{MobileNet search space results: comparison of the SOTA methods on ImageNet.} 
   \label{tab.5}
\end{table}

\subsection{Ablation Study and Analysis}
\label{sec.ablation.study}

We perform ablation study in this section. We analyze the impact of random labels and angle metric on RLNAS. All experiments are conducted on NAS-Bench-201. 

\paragraph{Methods of generating random labels.} 
In the above experiments, we uniformly sample random labels for images before SuperNet training and we denote it as (1). In this subsection, we further discuss 3 other methods for generating random labels:
(2).~shuffle all ground truth labels at once before SuperNet training,
(3).~uniformly sample labels in each training iteration, and
(4).~shuffle ground truth labels in each training iteration. According to these four methods, we conducted three repeated architecture search experiments across CIFAR-10, CIFAR-100 and ImageNet16-120.

As Table~\ref{tab.6} shows, in general, the methods of generating random labels at one time have higher performance than the methods of randomly generating labels in each iteration. Even if $\text{RLNAS}^{\dagger}$ has better performance than $\text{RLNAS}^{\ast}$ and $\text{RLNAS}^{\star}$ on CIFAR-10 and CIFAR-100, the performance on ImageNet16-120 is poor with a large margin and this means that $\text{RLNAS}^{\dagger}$ is instable and has poor transferring ability. As for $\text{RLNAS}^{\ast}$ and $\text{RLNAS}^{\star}$, these two methods obtain comparable test accuracy. Considering $\text{RLNAS}^{\ast}$ coupled with ground truth labels, we generate random labels with $\text{RLNAS}^{\star}$ in default and it is easy to apply our algorithm to tasks without labels.

\begin{table}[htbp]
   \centering 
   \footnotesize
   \begin{tabular}{c|c|c|c} 
   \hline 
   \multirow{2}*{Method} & CIFAR-10 & CIFAR-100  & ImageNet16-120 \\
   \cline{2-4}
      ~   & test acc~(\%) & test acc~(\%) & test acc~(\%) \\
   \hline\hline
   $\text{RLNAS}^\star$  & 93.45$\pm$0.11  & 70.71$\pm$0.36  & 43.70$\pm$1.25 \\
    \hline
   $\text{RLNAS}^\ast$ & 93.52$\pm$0.27 & 70.25$\pm$0.25 & \textbf{43.81}$\pm$\textbf{1.12}  \\
   \hline
   $\text{RLNAS}^\dagger$  & \textbf{93.65}$\pm$\textbf{0.07} & \textbf{71.45}$\pm$\textbf{0.42}  & 27.51$\pm$1.04 \\ 
   \hline
   $\text{RLNAS}^\ddagger$ & 92.85$\pm$0.46 & 61.59$\pm$6.57 & 27.51$\pm$1.04  \\
   \hline 
   \end{tabular}
   
   \caption{Search results of four generating random label method on NAS-Bench-201:~(1).$^\star$ uniform sample all random labels at once, (2).$^\ast$ shuffle all ground truth labels at once,
   (3).$^\dagger$ uniform sample labels in each iteration, and
   (4).$^\ddagger$shuffle ground truth labels in each iteration.} 
   \label{tab.6}
\end{table}

\begin{figure}[htbp]
   \centering
   \subfigure[Test accuracy]{
       \includegraphics[scale=0.27]{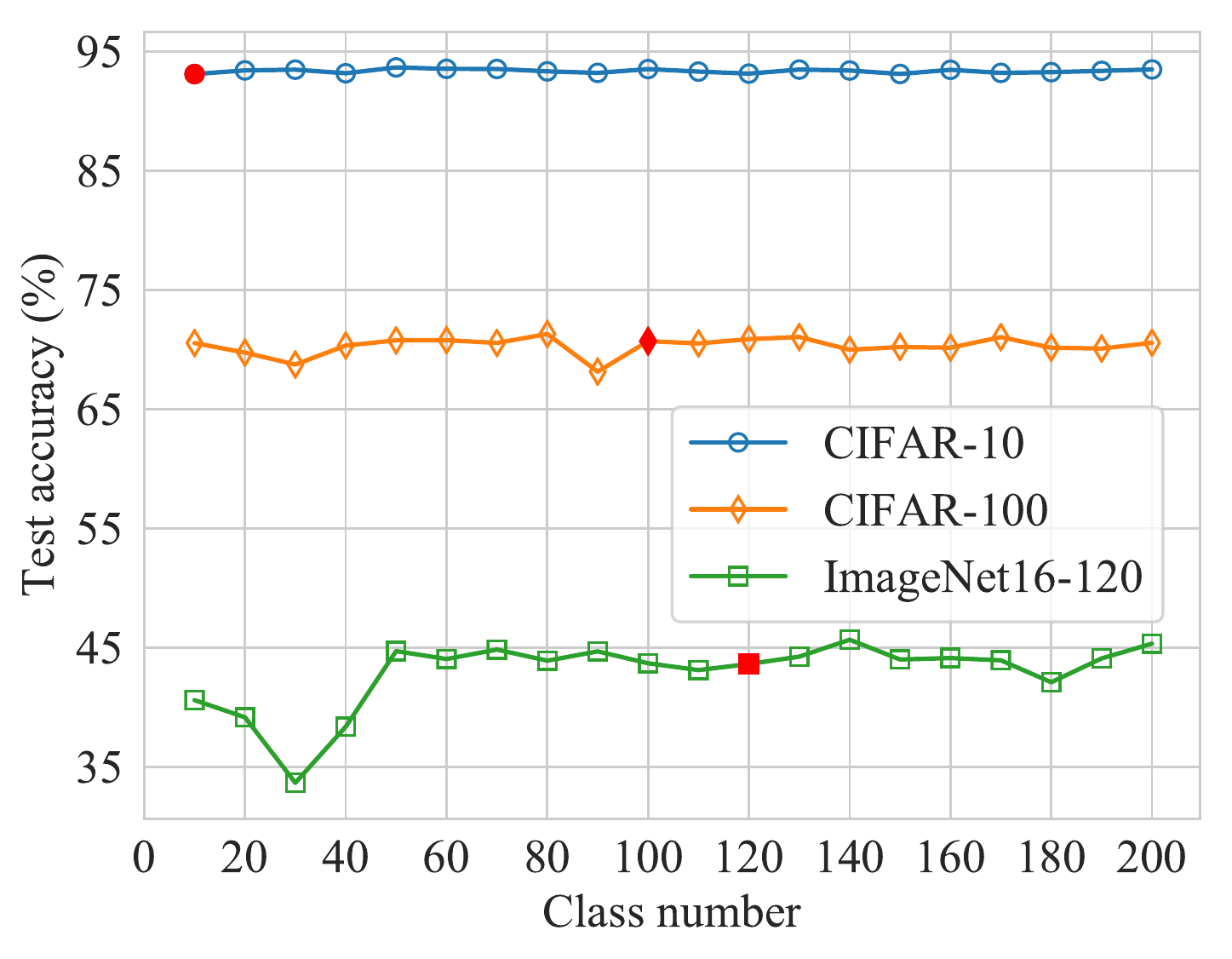}}
   \subfigure[Kendall's Tau]{
       \includegraphics[scale=0.27]{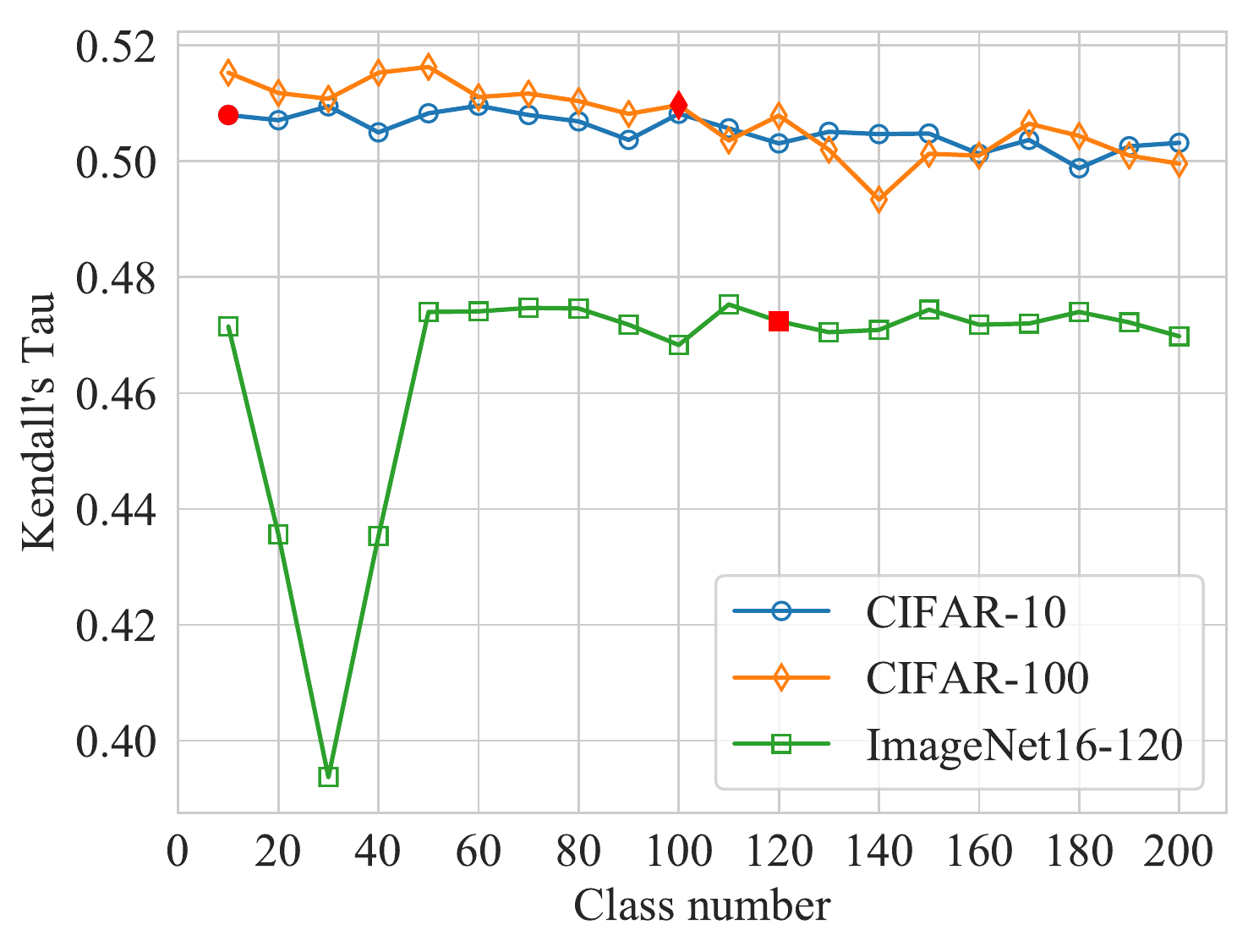}}
   \caption{Impact of the random label category on (a) test accuracy  and (b) Kendall's Tau~(best view in color). CIFAR-10, CIFAR-100 and ImageNet16-120 all sample 20 different image categories from 10 to 200 with interval 10. The red marker in each polyline represents the number of real image categories for different datasets.}
   \label{fig.1}
\end{figure}

\paragraph{Impact of image category.}
We have shown that uniform sample labels corresponding images before training is the most appropriate method to generate random labels. In this section, we further discuss the impact of the label category on searching performance. In detail, we sample 20 different categories from 10 to 200 with interval 10 for CIFAR-10, CIFAR-100 and ImageNet16-120. SuperNet is trained with different categories of random labels. After that, test accuracy and Kendall's Tau are obtained like subsection~\ref{sec.4.2.1}. As shown in Figure~\ref{fig.1}, test accuracy and Kendall's Tau fluctuate greatly when the number of categories on the ImageNet16-120 is small~(in $[10,50]$). However, Kendall's Tau and test accuracy are not sensitive to label categories in most cases. This observation implies that our method can be directly applied to tasks where the real image category is unknown.

\paragraph{Bias analysis of angle metric.} 
We have shown the impacts of random labels on RLNAS in the above section. Next, we further ablate the bias of angle metric in architecture search. Specifically, we initialize two SuperNet weights with the same distribution but different random seeds. Based on the SuperNet without training, evolution algorithm with angle is used to search architectures. We also construct a random search baseline which train SuperNet with uniform sampling strategy and ground truth labels, then randomly sample 100 architectures from NAS-Bench-201 search space. The top-1 architecture is selected among the sampled architectures according to their validation accuracy. Table~\ref{tab.7} compares our method with two training free methods with different initialization and one random search method. The results show that the two training free methods are worse than random search, and RLNAS is better than random search. This means that angle metric will not bias to a certain candidate architecture.

\begin{table}[htbp]
   \centering 
   \footnotesize
   \begin{tabular}{l|c|c|c} 
   \hline 
   \multirow{2}*{Method} & CIFAR-10 & CIFAR-100 & ImageNet16-120 \\
   \cline{2-4}
    ~  & test acc~(\%) & test acc~(\%) & test acc~(\%) \\
   \hline\hline
   $\text{Training free}^{\dagger}$  & 90.74$\pm$1.39 & 66.97$\pm$1.86 & 38.54$\pm$2.86 \\
   \hline
   $\text{Training free}^{\ddagger}$ & 91.55$\pm$1.34 & 66.59$\pm$2.10 & 39.03$\pm$3.91 \\
   \hline
   Random search & 92.09$\pm$0.21 & 67.27$\pm$1.28 & 40.77$\pm$3.64 \\
   \hline
   RLNAS & \textbf{93.45}$\pm$\textbf{0.11} & \textbf{70.71}$\pm$\textbf{0.36} & \textbf{43.70}$\pm$\textbf{1.25} \\  
   \hline 
   \end{tabular}
   
   \caption{Bias analysis of angle towards architectures on NAS-Bench-201.$^{\dagger}$ and $^{\ddagger}$ initializes model weights with normalization distribution and uniform distribution.} 
   \label{tab.7}
\end{table}

\subsection{Generalization Ability}

\begin{table*}[htbp]
   \centering 
   \footnotesize
   \begin{tabular}{l|c|c|c|c|c|c|c|c|c} 
   \hline 
   Method &  Params~(M)& FLOPs~(M) & Acc  & AP &$\text{AP}_{50}$ &$\text{AP}_{75}$ & $\text{AP}_{S}$ & $\text{AP}_{M}$ & $\text{AP}_{L}$\\
   \hline\hline
   Random search & 4.7 & 519 & 74.3 & 31.7 & 50.4 & 33.4 & 16.3 & 35.2 & 42.9 \\
   \hline
   DARTS-v1~\cite{liu2018darts}~(\textit{sup.})      & 4.5 & 507 & 74.3 & 31.2 & 49.5 & 32.6 & 16.1 & 33.9 & 43.6 \\
   DARTS-v2~\cite{liu2018darts}~(\textit{sup.})      & 4.7 & 531 & 74.9 & 31.5 & 50.3 & 33.1 & 16.9 & 34.5 & 43.0 \\
   P-DARTS~\cite{chen2019progressive}~(\textit{sup.})        & 4.9 & 544 & 75.7 & 32.9 & \textbf{52.1} & 34.7 & 17.2 & 36.2 & 44.8 \\
   PC-DARTS~\cite{xu2019pc}~(\textit{sup.})      & 5.3 & 582 & 75.9 & 32.9 & 51.8 & \textbf{34.8} & \textbf{17.5} & 36.3 & 43.5\\
   \hline
   UnNAS~\cite{liu2020labels}~(\textit{rotation task.})     & 5.1 & 552 & 75.8 & 32.8 & 51.5 & 34.7 & 16.7 & 36.1 & 44.5\\
   UnNAS~\cite{liu2020labels}~(\textit{color task.})   & 5.3 & 587 & 75.5 & 32.4 & 51.2 & 34.2 & 16.6 & 35.6 & 44.6\\
   UnNAS~\cite{liu2020labels}~(\textit{jigsaw task.})  & 5.2 & 560 & 76.2 & \textbf{33.0} & 51.9 & 35.3 & 16.4 & \textbf{37.2} & \textbf{45.4}\\
   \hline
   $\text{Ours}^\dagger$~(\textit{random label.})  & 5.5 & 597 & 75.9 & 32.4 & 50.9 & 34.4 & 16.5 & 35.5 & 44.5 \\
   $\text{Ours}^\ddagger$~(\textit{random label.})   & 5.2 & 561 & 75.9 & 32.9 & 51.6 & \textbf{34.8} & 16.8 & 36.7 & 44.5\\
   \hline
   \end{tabular}
   
   \caption{Object detection results of DARTS search space on MS COCO. $^\dagger$ search with 600M FLOPs constrain. $^\ddagger$ search without FLOPs constrain but scale FLOPs to 600M.} 
   \label{tab.8}
\end{table*}

\begin{table*}[htbp]
   \centering 
   \footnotesize
   \begin{tabular}{l|c|c|c|c|c|c|c|c|c} 
   \hline 
   Method &  Params~(M)& FLOPs~(M) & Acc  & AP &$\text{AP}_{50}$ &$\text{AP}_{75}$ & $\text{AP}_{S}$ & $\text{AP}_{M}$ & $\text{AP}_{L}$\\
   \hline\hline
   Random search~(\textit{sup.}) & 4.5 & 446 & 75.3 & 29.7 & 47.5 & 31.4 & 15.3 & 32.6 & 39.9 \\
   \hline
   FairNAS-A~\cite{chu2019fairnas}~(\textit{sup.})     & 4.7 & 389 & 75.1 & 29.8 & 47.8 & 31.4 & 15.5 & 32.3 & \textbf{41.2} \\
   Proxyless~(GPU)~\cite{cai2018proxylessnas}~(\textit{sup.})  & 7.0 & 457 & 75.5 & 29.5 & 47.5 & 30.9 & 15.5 & 32.4 & 40.8 \\
   FairDARTS-D~\cite{chu2019fair}~(\textit{sup.})   & 4.4 & 477 & 74.7 & 29.6 & 47.2 & 31.1 & 14.6 & 32.5 & 40.1\\
   SPOS~\cite{guo2019single}~(\textit{sup.})          & 5.4 & 472 & 75.6 & 29.8 & \textbf{48.1} & 31.1 & \textbf{16.0} & 32.6 & 40.4 \\
   \hline
   Ours~(\textit{unsup.})          & 5.3 & 473 & 75.6 & \textbf{30.0} & 47.6 & \textbf{31.8} & 15.7 & \textbf{32.8} & 40.5\\
   \hline
   \end{tabular}
   
   \caption{Object detection results of MobileNet-like search space on MS COCO.} 
   \label{tab.9}
\end{table*}

We evaluate the generalization ability of RLNAS on two downstream tasks:~object detection and semantic segmentation. We first retrain the models searched by different NAS methods on ImageNet , and then finetune these pre-trained models on downstream tasks. In order to make fair comparisons, models searched in the same search space adopt the same training setting for ImageNet classification tasks. At the same time, models for the same downstream task also use the same training setting, no matter what search space the model is searched from.

\paragraph{Object detection.} We conduct experiments on MS COCO~\cite{Lin2014Microsoft} and adopt RetinaNet~\cite{2017Focal} as the detection framework. The train and test image scale is 800$\times$ resolution. We only modify the backbone of RetinaNet and train RetinaNet with default training setting as Detectron2~\cite{wu2019detectron2}. Table~\ref{tab.8} and Table~\ref{tab.9} show the comparisons of models searched in DARTS and MobileNet-like search space respectively. RLNAS obtains comparable AP in DARTS search space and surpasses other methods with slight margin in MobileNet-like search space.

\paragraph{Semantic segmentation.} We further test RLNAS on the task of semantic segmentation on Cityscapes~\cite{2016The} dataset. We adopt DeepLab-v3~\cite{chen2017rethinking} as segmentation framework. The train and test image scale is 769$\times$769 and we train DeepLab-v3 with 40k iterations. The other segmentation training setting are kept the same as MMSegmentation~\cite{mmseg2020}. Table~\ref{tab.10} and Table~\ref{tab.11} make comparisons among models searched on DARTS and MobileNet-like search space respectively. For DARTS search space, $\text{RLNAS}^{\dagger}$ obtains 73.2\% mIoU and outperform other methods by a large margin. RLNAS also obtains comparable mIoU compared to other methods in MobileNet search space.

\paragraph{Summary.} We conclude that RLNAS achieves comparable  or even superior performance across two downstream tasks and various search spaces, without bells and whistles. 

\begin{table}[htbp]
   \centering 
   \footnotesize
   \begin{tabular}{l|c|c|c|c} 
   \hline 
   Method &   \makecell[c]{Params \\ (M)}&  \makecell[c]{FLOPs \\ (M)} &  \makecell[c]{Acc \\ (\%)} &  \makecell[c]{mIoU \\ (\%)} \\   \hline\hline
   Random search~(\textit{sup.}) & 4.7 & 519 & 74.3 & 72.3 \\
   \hline
   DARTS-v1~\cite{liu2018darts}~(\textit{sup.})      & 4.5 & 507 & 74.3 & 72.7 \\
   DARTS-v2~\cite{liu2018darts}~(\textit{sup.})      & 4.7 & 531 & 74.9 & 71.8 \\
   P-DARTS~\cite{xu2019pc}~(\textit{sup.})      & 4.9 & 544 & 75.7 & 71.9 \\
   PC-DARTS~\cite{xu2019pc}~(\textit{sup.})      & 5.3 & 582 & 75.9 & 72.2 \\
   \hline
   UnNAS~\cite{liu2020labels}~(\textit{rotation task.})      & 5.1 & 552 & 75.8 & 71.9 \\
   UnNAS~\cite{liu2020labels}~(\textit{color task.})    & 5.3 & 587 & 75.5 & 72.0 \\
   UnNAS~\cite{liu2020labels}~(\textit{jigsaw task.})   & 5.2 & 560 & 76.2 & 72.1 \\
   \hline
   $\text{Ours}^\dagger$~(\textit{random label.})    & 5.5 & 597 & 75.9 & \textbf{73.2} \\
   $\text{Ours}^\ddagger$~(\textit{random label.})   & 5.2 & 561 & 75.9 & 72.5 \\
   \hline

   \end{tabular}
   \caption{Semantic segmentation results of DARTS search space on Cityscapes.$^\dagger$ search with 600M FLOPs constrain. $^\ddagger$ search without FLOPs constrain but scale FLOPs to 600M.} 
   \label{tab.10}
\end{table}

\begin{table}[htbp]
   \centering 
   \footnotesize
   \begin{tabular}{l|c|c|c|c} 
   \hline 
   Method &   \makecell[c]{Params \\ (M)}&  \makecell[c]{FLOPs \\ (M)} &  \makecell[c]{Acc \\ (\%)} &  \makecell[c]{mIoU \\ (\%)} \\
   \hline\hline
   Random search~(\textit{sup.}) & 4.5 & 446 & 75.3 & 70.6 \\
   \hline
   FairNAS-A~\cite{chu2019fairnas}~(\textit{sup.})     & 4.7 & 389 & 75.1 & 72.0 \\
   Proxyless~(GPU)~\cite{cai2018proxylessnas}~(\textit{sup.})  & 7.0 & 457 & 75.5 & 71.0 \\
   FairDARTS-D~\cite{chu2019fair}~(\textit{sup.})   & 4.4 & 477 & 74.7 & \textbf{72.1} \\
   SPOS~\cite{guo2019single}~(\textit{sup.})          & 5.4 & 472 & 75.6 & 71.6 \\
   \hline
   Ours~(\textit{unsup.})         & 5.3 & 473 & 75.6 & 71.8 \\
   \hline
   \end{tabular}
   
   \caption{Semantic segmentation results of MobileNet-like search space on Cityscapes.} 
   \label{tab.11}
\end{table}

\newpage

\newpage

\newpage

\bibliographystyle{ieee_fullname}
\bibliography{reference}

\clearpage
\section{Appendix}
\subsection{Architectures Searched in DARTS Search Space}
\begin{figure}[htbp]
   \centering
   \subfigure[normal cell]{
       \includegraphics[scale=0.25]{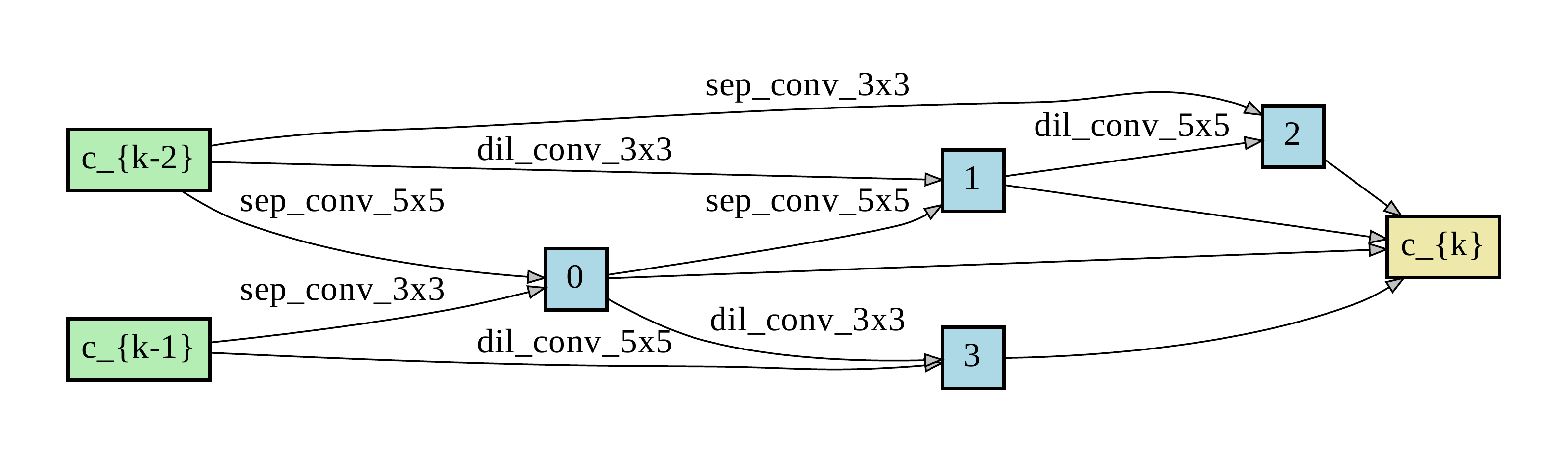}}
   \subfigure[reduce cell]{
       \includegraphics[scale=0.25]{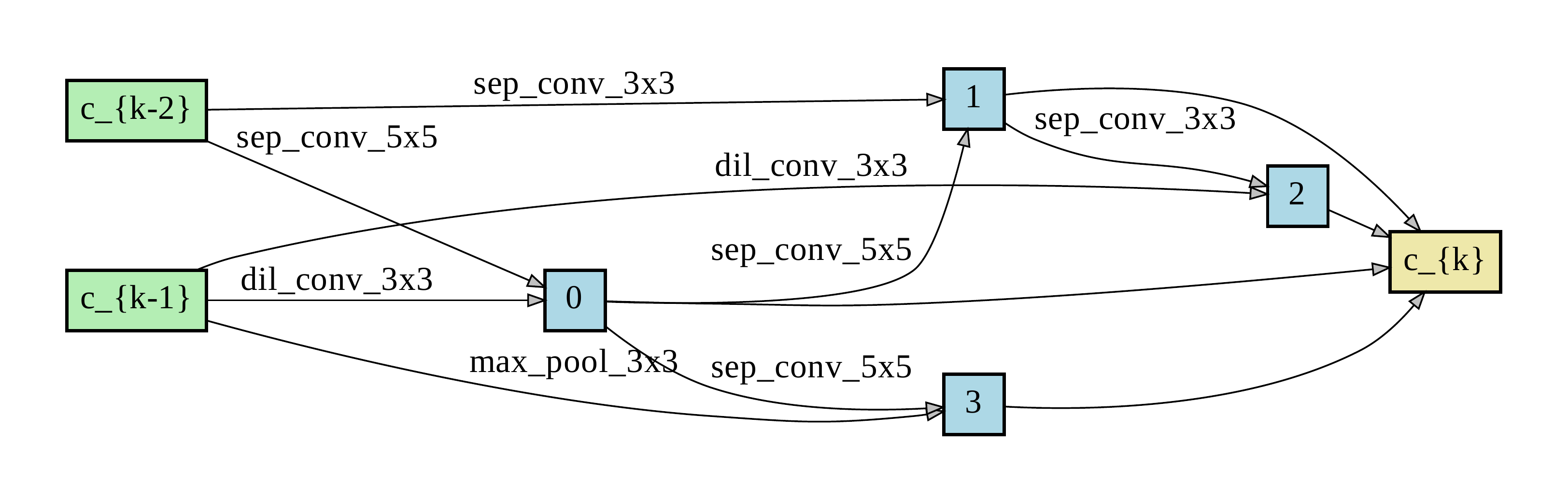}}
   \caption{The best architecture of RLNAS searched on CIFAR-10 dataset.}
   \label{fig.2}
\end{figure}

\begin{figure}[htbp]
   \centering
   \subfigure[normal cell]{
       \includegraphics[scale=0.25]{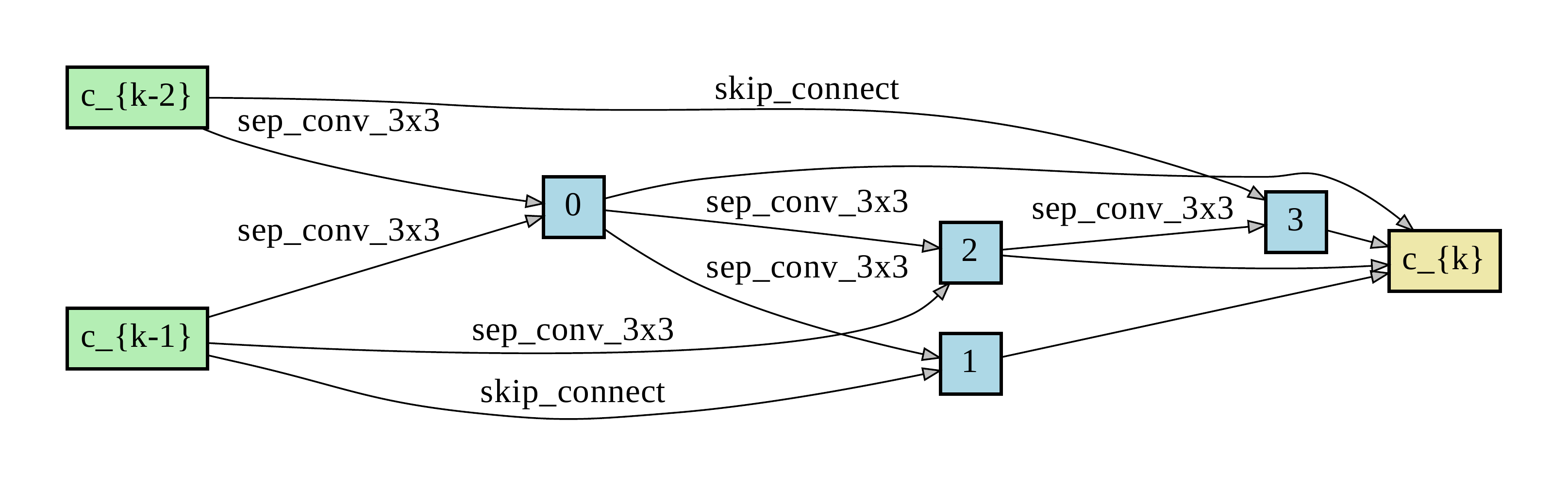}}
   \subfigure[reduce cell]{
       \includegraphics[scale=0.25]{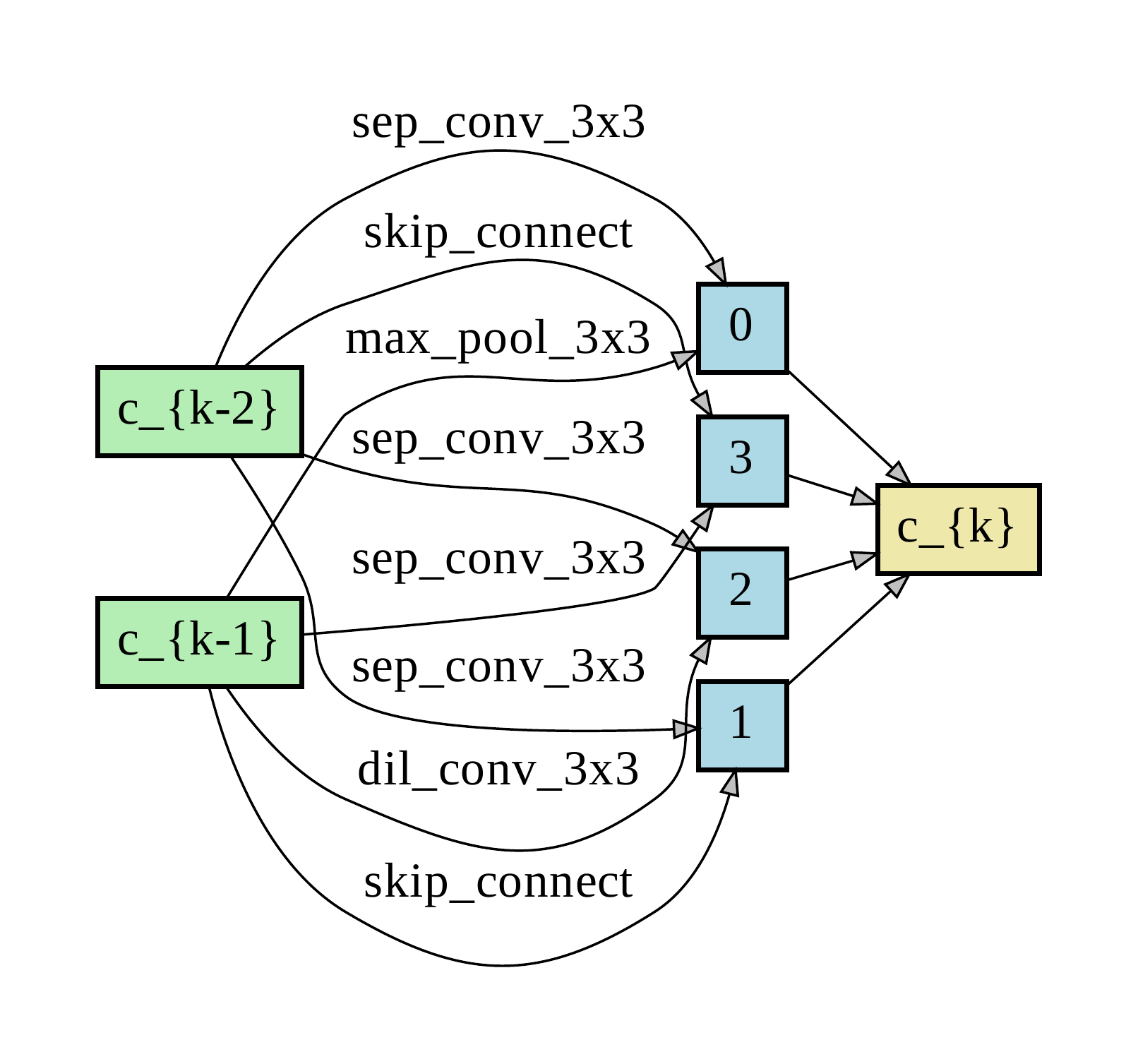}}
   \caption{The best Architecture of RLNAS searched on ImageNet dataset within 600M FLOPs constrain.}
   \label{fig.3}
\end{figure}

\begin{figure}[htbp]
   \centering
   \subfigure[normal cell]{
       \includegraphics[scale=0.25]{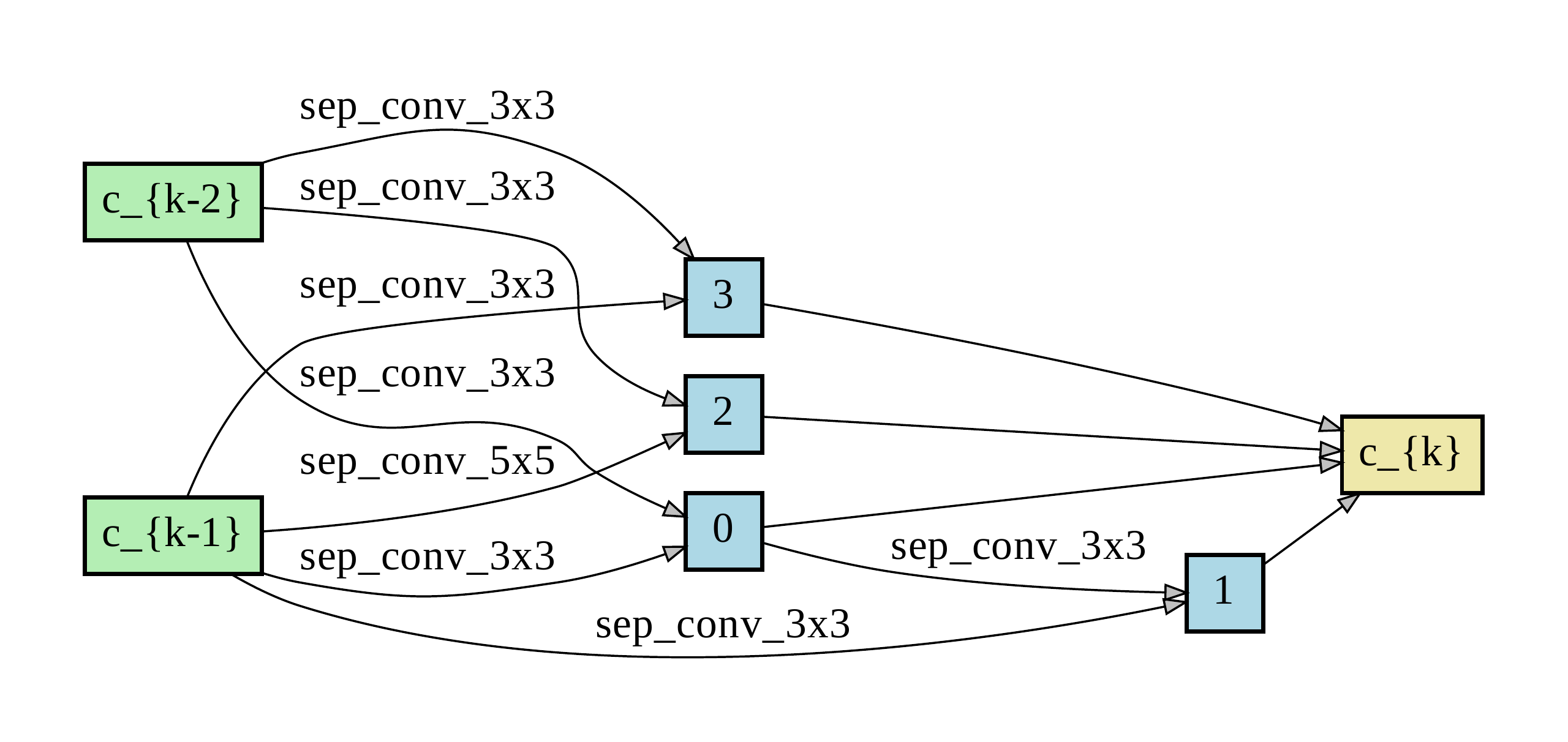}}
   \subfigure[reduce cell]{
       \includegraphics[scale=0.25]{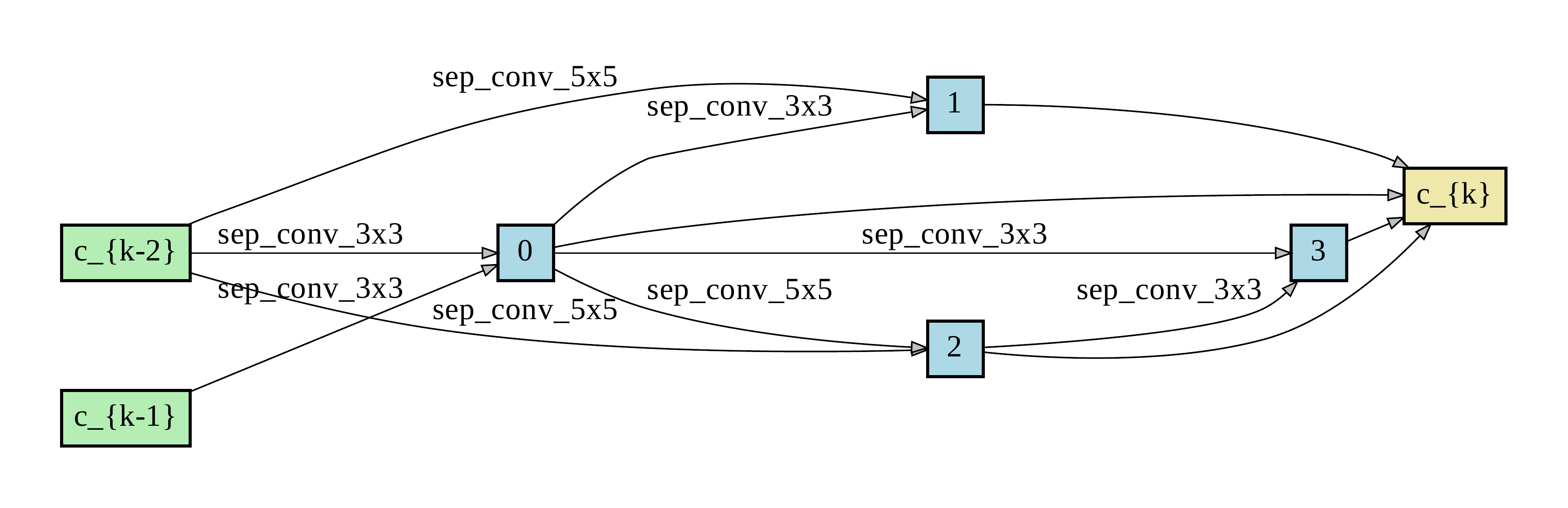}}
   \caption{The best architecture of RLNAS searched on ImageNet dataset without FLOPs constrain.}
   \label{fig.4}
\end{figure}

In DARTS search space, we visualize all RLNAS architectures :~searched on CIFAR-10~(Figure~\ref{fig.2}), ImageNet within 600M FLOPs constrain~(Figure~\ref{fig.3}), ImageNet without Flops constrain~(Figure~\ref{fig.4}). 

\subsection{Architectures Searched in MobileNet-like Search Space}
\begin{figure}[htbp]
   \centering
   \includegraphics[scale=0.25]{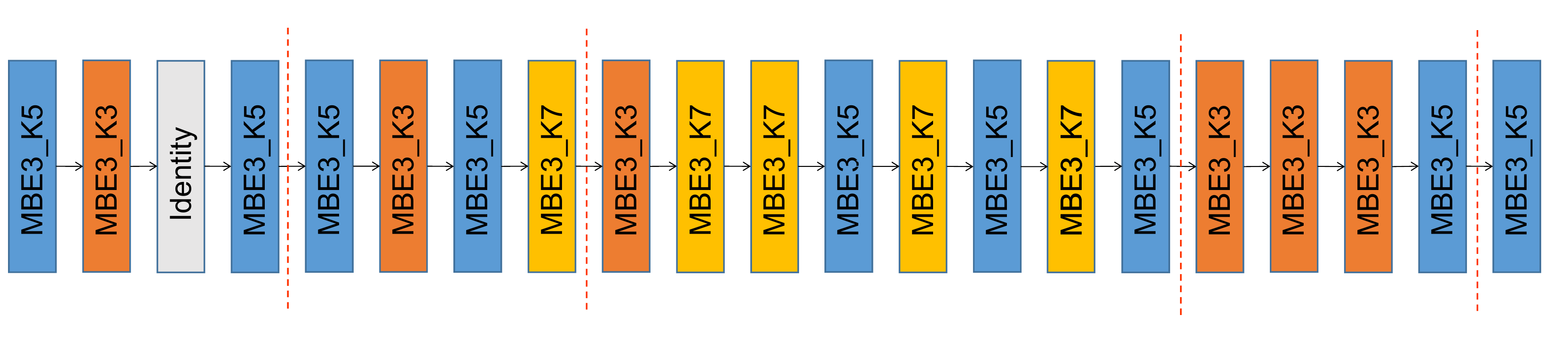}
   \caption{The best architecture of RLNAS searched on ImageNet dataset within 475M FLOPs constrain.}
   \label{fig.5}
\end{figure}

In MobileNet-like search space, we visualize the architecture searched on ImageNet~(Figure~\ref{fig.5}).

\subsection{Comparison with UnNAS on NAS-Bench-201}
We further conduct experiments on NA-Bench-201 to compare with UnNAS. We use the same pretext tasks on CIFAR-10 as UnNAS. Specifically, we leverage SPOS with pretext tasks to train supernet and the validation accuracy of pretext tasks is used as fitness to evolve architecture search. As Table~\ref{tab.12} shows, RLNAS obtains architectures with higher test accuracy but lower accuracy variance.

\begin{table}[htbp]
   \centering 
   \footnotesize
   \begin{tabular}{l|c} 
   \hline 
   \multirow{2}*{Method} & CIFAR-10 \\
   \cline{2-2}
    ~  & test acc~(\%)  \\
   \hline\hline
   UnNAS~\cite{liu2020labels}~(\textit{rotation task}) & 92.41$\pm$0.12  \\
   \hline
   UnNAS~\cite{liu2020labels}~(\textit{color task}) & 92.14$\pm$0.60 \\
   \hline
   UnNAS~\cite{liu2020labels}~(\textit{jigsaw task}) & 92.38$\pm$0.19  \\
   \hline
   RLNAS & \textbf{93.45}$\pm$\textbf{0.11}  \\
   \hline 
   \end{tabular}
   
   \caption{Comparison with UnNAS on NAS-Bench-201.} 
   \label{tab.12}
\end{table}

\end{document}